\documentclass{article}

\usepackage{booktabs,threeparttable,multirow,amssymb,tabularx,array,amsmath,url,caption,changepage,cite,float,multirow,tikz}
\usepackage{authblk} 
\usepackage[linesnumbered,ruled,vlined]{algorithm2e}
\usepackage[utf8]{inputenc}
\usepackage{arxiv}

\def\checkmark{\tikz\fill[scale=0.4](0,.35) -- (.25,0) -- (1,.7) -- (.25,.15) -- cycle;}

\title{Zero-shot 3D Map Generation with LLM Agents: A Dual-Agent Architecture for Procedural Content Generation}

\author[1,*]{Lim Chien Her} 
\author[1]{Ming Yan}
\author[1]{Yunshu Bai}
\author[1]{Ruihao Li}
\author[1]{Hao Zhang}

\affil[1]{MiAO}

\date{}

\begin{document}
\maketitle

\renewcommand{\thefootnote}{\fnsymbol{footnote}}
\footnotetext[1]{Correspondence to: limchienher@miao.company}

\begin{figure}[H]
    \centering
    \includegraphics[width=15cm]{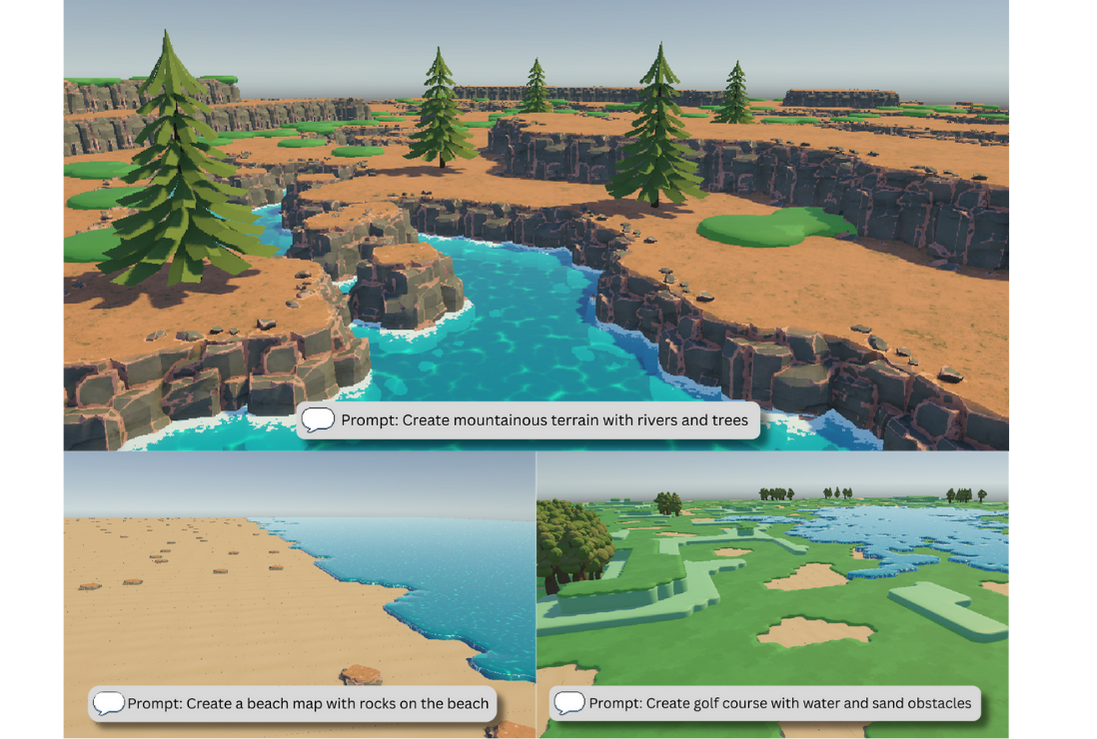}
    \caption{Our proposed training-free architecture generates realistic 3D maps from natural language prompts using procedural content generation in a zero-shot manner.}
    \label{fig:IntroductionImage}
\end{figure}
\begin{abstract}
\noindent Procedural Content Generation (PCG) offers scalable methods for algorithmically creating complex, customizable worlds. However, controlling these pipelines requires the precise configuration of opaque technical parameters. We propose a training-free architecture that utilizes LLM agents for zero-shot PCG parameter configuration. While Large Language Models (LLMs) promise a natural language interface for PCG tools, off-the-shelf models often fail to bridge the semantic gap between abstract user instructions and strict parameter specifications. Our system pairs an Actor agent with a Critic agent, enabling an iterative workflow where the system autonomously reasons over tool parameters and refines configurations to progressively align with human design preferences. We validate this approach on the generation of various 3D maps, establishing a new benchmark for instruction-following in PCG. Experiments demonstrate that our approach outperforms single-agent baselines, producing diverse and structurally valid environments from natural language descriptions. These results demonstrate that off-the-shelf LLMs can be effectively repurposed as generalized agents for arbitrary PCG tools. By shifting the burden from model training to architectural reasoning, our method offers a scalable framework for mastering complex software without task-specific fine-tuning.
\end{abstract}
\section{Introduction}\label{1}
Procedural Content Generation (PCG) plays a critical role in modern game development by allowing the scalable creation of sufficient and varied content for contemporary virtual environments, thus significantly reducing manual production requirements \cite{mohaghegh2023pcgptproceduralcontentgeneration,Todd_2023,algorithm_selection}. PCG operates through generative algorithms that are parameterized and often combined in multi-layered pipelines. The process typically begins with a seed value, ensuring deterministic output, and applies a series of rules to transform basic geometric or semantic primitives into large-scale, internally consistent game environments \cite{traditional_PCG}. By leveraging these predefined rules and algorithms, PCG supports the scalable creation of varied and configurable game environments. Variants of this approach are employed in commercial titles such as \textit{Minecraft} \cite{mojang2011minecraft}, \textit{Diablo} \cite{diablo} and \textit{Valheim} \cite{irongate2021valheim}.
\\\\
Despite widespread adoption \cite{PCG_book}, authoring effective PCG pipelines remains a challenging task. The artistic and functional qualities of the output are controlled by a dense set of technical parameters that are often opaque, non-intuitive, and lack a direct mapping to high-level design concepts. As such, navigating these high-dimensional parameter spaces \cite{5756645} and selecting appropriate algorithms \cite{algorithm_selection} require substantial expertise and often involve iterative trial and error \cite{parameter_trial_error}.
\\\\
Recent developments in large language models (LLMs) have prompted research into using natural language to control these pipelines\cite{sun2023languagerealitycocreativestorytelling,Zhu_2023,sudhakaran2023mariogptopenendedtext2levelgeneration}. However, off-the-shelf LLMs struggle to bridge the semantic gap between abstract user instructions and strict parameter specifications \cite{LLMsarebadatParameters}. While adept at linguistic interpretation \cite{LLMsAreGoodatLanguage}, standard models lack the grounding to assign functionally appropriate values to domain-specific variables, often resorting to speculative guessing based on parameter names. To mitigate this, current approaches rely on extensive fine-tuning \cite{nasir2023practicalpcglargelanguage}, reinforcement learning \cite{baek2024chatpcglargelanguagemodeldriven}, or in-domain pre-training \cite{pretraining_importance}. These methods are resource-intensive \cite{lora,RLHF} and result in rigid models that lack generalizability across different tools or domains \cite{dontstoppretrainingadapt,modelsoupsaveragingweights}.
\begin{figure}[t]
    \centering
    \includegraphics[width=15cm]{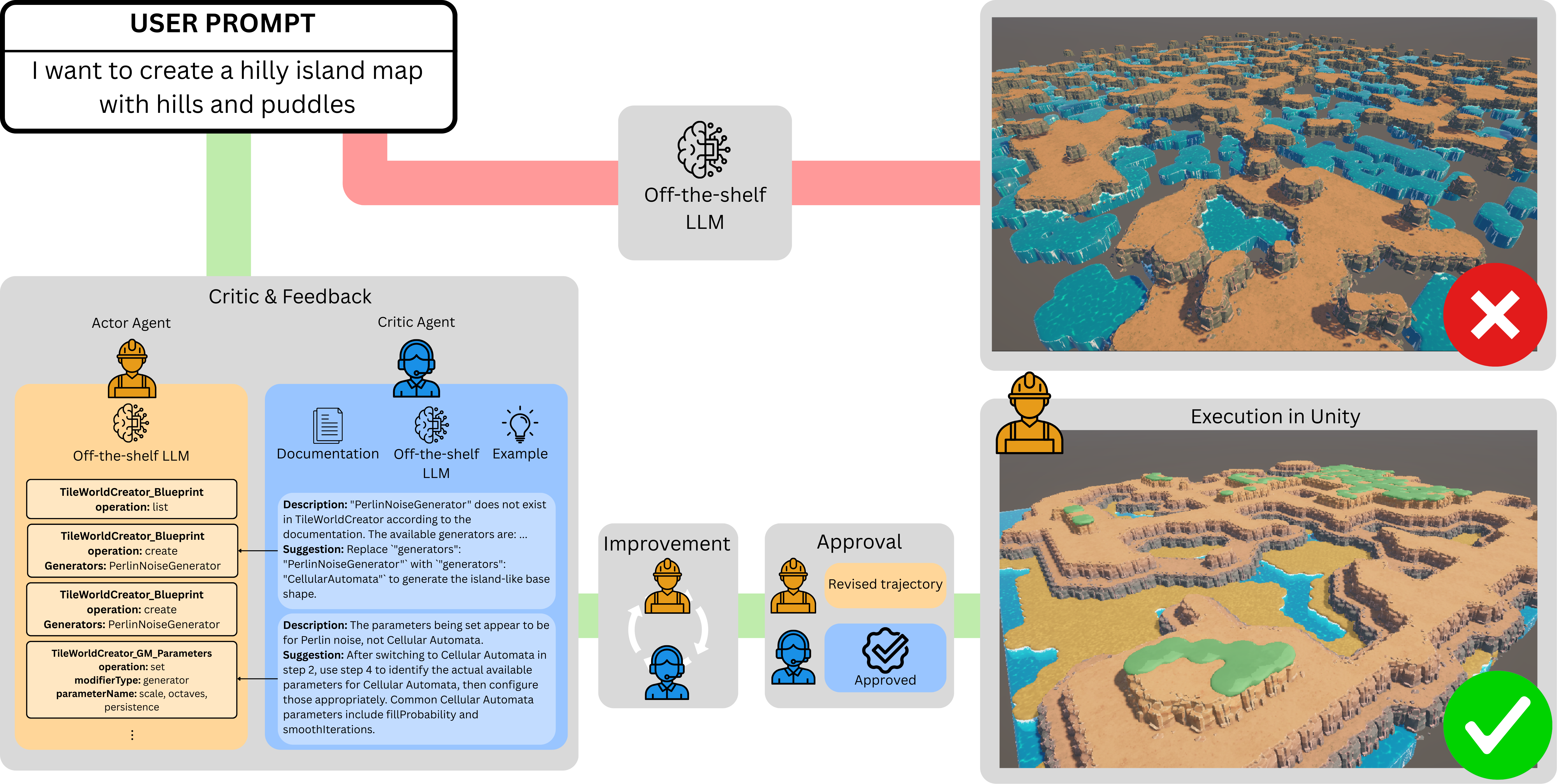}
    \caption{Our proposed architecture employs the use of two agents which interact dialogically to achieve understanding of opaque and non-intuitive parameters of PCG tools. Training-free and in zero-shot, the architecture is able to produce complex 3D maps.}
    \label{fig:Workflow_Comparison}
\end{figure}To address these limitations, we introduce a training-free, dual-agent architecture that enables off-the-shelf LLMs to interface with advanced PCG tools in a zero-shot setting, presented in Figure~\ref{fig:Workflow_Comparison}. Our system pairs an Actor agent, which translates natural language prompts into initial parameter proposals, with a Critic agent, which evaluates these proposals against algorithmic documentation and constraints. Through a dialogic process, the agents iteratively refine the configuration, allowing the system to resolve ambiguities and correct errors autonomously. This interaction replaces the need for gradient updates with in-context reasoning, ensuring that valid parameters are selected based on logic rather than memorized patterns.
\\\\
Critically, this approach demonstrates that general-purpose LLMs can be repurposed as flexible, domain-agnostic controllers. By shifting the burden from model training to architectural design, our method requires no task-specific fine-tuning or data collection. This allows for immediate transfer to other new tools simply by swapping the provided reference documentation.
\\\\
Our primary contributions are summarized as follows:
\begin{enumerate}
    \item \textbf{A Training-Free Framework for Zero-Shot PCG:} We demonstrate that off-the-shelf LLMs can operate complex, domain-specific software solely through the injection of static documentation and reference examples, eliminating the need for resource-intensive fine-tuning or reinforcement learning.
    \item \textbf{Dual-Agent Actor–Critic Architecture:} We introduce a novel architecture that decouples \textit{semantic planning} (Actor) from \textit{functional verification} (Critic). We show that this dialogic separation of concerns allows the system to autonomously resolve ambiguities and correct parameter hallucinations without human intervention.
    \item \textbf{Empirical Validation of Reliability:} Through quantitative experiments on 2D and 3D map generation tasks, we establish that our dual-agent approach achieves a 20\% higher task success rate compared to single-agent baselines, effectively bridging the semantic gap between natural language and high-dimensional parameter spaces.
\end{enumerate}
\section{Related Work}\label{sec2}
\paragraph{Traditional PCG.}
Procedural Content Generation (PCG) refers to the algorithmic creation of game content \cite{PCG_book}. Since the 1980s, PCG has evolved into three main paradigms: search-based, machine learning (ML)-based, and large language model (LLM)-based approaches \cite{PCG_survey,PCG_book,traditional_PCG}. Search-based PCG typically utilizes evolutionary algorithms, where content is evolved via a fitness function that assesses quality against specific criteria \cite{Search_based_PCG,10.1145/3582437.3587196}. While effective for optimizing well-defined constraints, these methods rely heavily on manually designed fitness functions and lack the semantic flexibility to interpret vague or high-level natural language instructions, limiting their utility in rapid, intent-driven design workflows. Machine Learning (ML)-based approaches employ models such as Generative Adversarial Networks (GANs), Long Short-Term Memory (LSTM) and Autoencoders to learn generative patterns from existing game corpora \cite{PCG_survey}. While these methods can capture complex stylistic nuances that rule-based systems miss, they suffer from a 'black box' problem: the resulting content is often difficult to control or edit post-generation \cite{MLPCG}. Furthermore, training these models requires large annotated datasets of game levels, which are rarely available \cite{MLPCG}. This data dependency makes ML-based approaches brittle when applied to new domains without extensive retraining.

\paragraph{LLM-PCG.}
To incorporate natural language control, prevailing methods rely on fine-tuning or reinforcement learning (RL) to equip models with map generation capabilities. For instance, Muhammad and Togelius \cite{nasir2023practicalpcglargelanguage} finetuned GPT-3 using human-edited data of room levels generated by LLM. Whereas ChatPCG \cite{baek2024chatpcglargelanguagemodeldriven} performed Reinforcement Learning with Artificial Intelligence feedback (RLAIF) to align a language model to produce playable maps using PCG. However, these approaches incur significant overhead: fine-tuning requires large quantities of high-quality and domain-specific data \cite{szep2025finetuninglargelanguagemodels,nasir2023practicalpcglargelanguage,Brown2020},which can be costly to produce or curate, while RLAIF depends on a stronger critic model and requires training of a reward model \cite{sharma2024criticalevaluationaifeedback,baek2024chatpcglargelanguagemodeldriven}. As a result, these methods are difficult to generalize or repurpose when tools, genres, or design goals change. Off-the-shelf LLM implementations that avoid fine-tuning typically position the model as a natural-language-to-asset bridge rather than as a direct operator of PCG algorithms. Systems like Auxtero \cite{auxtero2023game} and Chen et al. \cite{chen2025narrativetoscenegenerationllmdrivenpipeline} decomposed prompts into object lists for tile-based placement. Crucially, existing research predominantly targets 2D environments, despite modern games prioritizing 3D design \cite{PCG_survey}. Current 2D work focuses on top-down map or obstacle generation (e.g., Sokoban levels \cite{nasir2023practicalpcglargelanguage,mohaghegh2023pcgptproceduralcontentgeneration,Todd_2023}), leaving 3D terrain synthesis underexplored. These pipeline architectures successfully translate user prompts into executable code or asset selection steps, but often sidestep the hardest part of PCG tool use: choosing and configuring generation algorithms via dense, opaque parameter sets. LLMs are restricted to issuing high-level commands, with the core generative logic and parameterization delegated to hand-engineered components.

\paragraph{In-Context Learning and Tool Use.} Unlike traditional ML approaches that require parameter updates, Large Language Models demonstrate the emergent ability to perform tasks via In-Context Learning (ICL), where the model infers rules and patterns directly from the prompt \cite{Brown2020}. Recent work has extended this capability to "Tool Use," where LLMs function as controllers for external APIs and software interfaces by interpreting documentation and generating executable syntax zero-shot \cite{toolformer,Mialon2023,ReAct}. This paradigm shifts the engineering burden from dataset curation to prompt design. However, most existing research in tool use focuses on basic retrieval or calculator APIs. Our work extends this by investigating whether ICL is sufficient for mastering the high-dimensional, inter-dependent parameter spaces found in professional 3D content creation tools.

\paragraph{Multi-agent Architectures.}
Recent multi-agent architectures retain this controller-centric paradigm by tethering LLMs to external game engines or modeling software, despite demonstrating improved fidelity in 3D scene generation. For example, UnrealLLM \cite{2025-unrealllm} employs agents to translate natural language into Unreal Engine 5’s PCG blueprints via a constructed knowledge base, effectively delegating terrain synthesis to the engine. SceneCraft \cite{hu2024scenecraftllmagentsynthesizing} and LL3M \cite{lu2025ll3mlargelanguage3d} generate editable Blender code through planning, retrieval, and coding agents to improve robustness and error handling. 3D-GPT \cite{taskdecompose_llm} adopts a three-agent design in which one agent identifies key functional elements in map specifications, another resolves ambiguities by elaborating vague descriptions, and a third produces Blender API function calls. While these systems show that multi-agent coordination can enhance reliability and expressivity, they rely on substantial bespoke scaffolding—such as curated knowledge bases, retrieval pipelines, or API-specific adapters. This infrastructure is time-consuming to construct and difficult to transfer to new, high-dimensional parameter spaces common in commercial tools. In contrast, our work enables LLM agents to master such tools through purely in-context learning. By utilizing task demonstrations to guide parameter synthesis without weight updates or bespoke adapters, we democratize LLM-driven PCG for complex 3D terrain generation.
\section{Method}
\begin{figure}[!ht]
    \centering
    \includegraphics[width=0.8\textwidth]{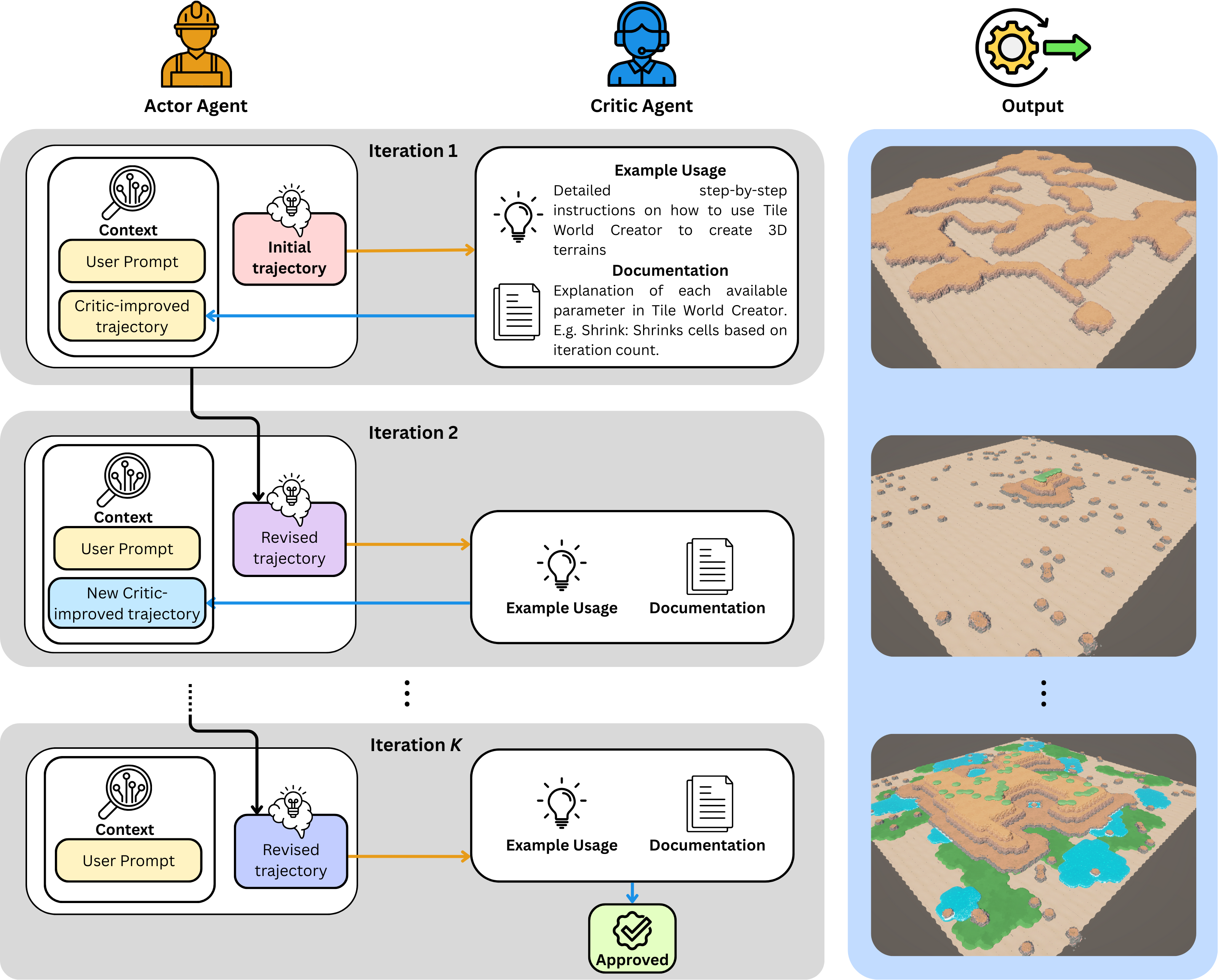}
    \caption{The Actor and Critic interacts dialogically to continuously refine the trajectory until (1) the Critic approves or (2) maximum iterations has been reached. Following which, the Actor creates the map based on the latest trajectory.}
    \label{fig:IterativeRefinement}
\end{figure}
\noindent We introduce a dual-agent architecture to support zero-shot adaptation of large language models to PCG tools and to improve their ability to interpret and assign values to domain-specific parameters. This framework addresses the challenge of interpreting high-dimensional domain parameters without fine-tuning. The architecture comprises two LLM-based agents, an Actor and a Critic (section 3.1) , functioning within an iterative dialogic loop (section 3.3).
\subsection{Dual Agent Architecture}
\paragraph{Actor.} The Actor functions as the semantic interpreter and generator. Given a user’s natural language prompt ($P_{user}$) describing the desired map, the Actor parses the intent and synthesizes an initial Parameter Trajectory Sequence ($S_0$). This sequence maps high-level, open-ended design concepts into specific PCG algorithm steps and their corresponding low-level parameter values. By operationalizing abstract descriptors into concrete tool settings, the Actor bridges the semantic gap between human intent and rigid software interfaces.

\paragraph{Critic.} The generated trajectory ($S_0$) is passed to the Critic agent, which acts as a static verifier. The Critic evaluates the proposed parameters against two provided knowledge sources:
\begin{enumerate}
    \item \textbf{API Documentation:} Formal definitions of available PCG algorithms and parameter constraints.
        \item \textbf{Reference Demonstration:} Validated example of algorithm orchestration for 3D map generation.
\end{enumerate}
Unlike the Actor, which prioritizes semantic alignment, the Critic prioritizes functional correctness. It identifies discrepancies where syntactically valid parameters may be operationally flawed (e.g., out-of-bounds values or incompatible algorithm pairings). As presented in Figure ~\ref{fig:Workflow_Comparison}, for every detected error, the Critic outputs a structured critique containing a description of the error and a correction suggestion, grounding the generation in authoritative specifications.

\begin{table}[h]
\centering
\fontsize{9pt}{9pt}\selectfont
\caption{Operational Comparison of Actor and Critic Agents}
\label{tab:agent_comparison}
\begin{tabularx}{\textwidth}{@{}l l X X X@{}}
\toprule
\textbf{Agent} & \textbf{Role} & \textbf{Operational Priority} & \textbf{Rules and Guidelines} & \textbf{Output Structure} \\
\midrule
\textbf{Actor} & 
  Semantic Interpreter &
  \textbf{Semantic Alignment}: Bridges the gap between abstract user intent and rigid tool settings. &
  \textbf{No Immediate Execution}: Forbidden from executing tools directly. \newline
  \textbf{Risk Assessment}: Must explicitly list assumptions or potential risks. &
  \textbf{Parameter Trajectory} \newline
  \texttt{\{trajectory\_summary, tool\_plan, risks\}} \\
\addlinespace[10pt]
\textbf{Critic} & 
  Static Verifier &
  \textbf{Functional Correctness}: Prioritizes validity against API documentation over creativity. &
  \textbf{Conservative Certainty}: Only flag ``blocking issues'' if absolutely certain. \newline
  \textbf{Contextual Review}: Verify against docs and usage examples. &
  \textbf{Structured Critique} \newline
  \texttt{\{decision, blocking\_issues, correction\_suggestion\}} \\
\bottomrule
\end{tabularx}
\end{table}

\subsection{Prompt Formulation}
To facilitate zero-shot adaptation, we employ a structured prompt engineering strategy that grounds both agents in domain documentation while enforcing strict output formats.
\\\\
\textbf{Actor Prompt Structure.} The Actor is initialized with a system instruction establishing its persona as a Unity tool planner. The prompt explicitly decouples planning from execution, strictly forbidding the agent from executing tools itself. Instead, it is tasked with proposing a concrete Execution Trajectory. The Actor is constrained to output a strict JSON structure comprising:
\begin{enumerate}
    \item \textbf{Trajectory Summary:} A high-level overview of the plan.
    \item \textbf{Tool Plan:} A granular list of sequential steps, where each step defines an \texttt{objective}, the specific \texttt{tool\_name}, the required \texttt{arguments}, and an \texttt{expected\_result} for verification.
    \item \textbf{Risk Assessment:} A dedicated field for identifying potential blocking risks or missing information before the plan is submitted
\end{enumerate}
\textbf{Critic Prompt Structure.} The Critic functions as a Plugin Expert tasked with validating the Actor's proposed trajectory against provided documentation and usage examples. Its system prompt enforces a comprehensive Review Framework covering five distinct dimensions: Tool Selection, Parameter Correctness, Logic \& Sequence, Goal Alignment, and a Certainty Requirement. Crucially, the Critic is instructed to adopt a conservative review policy: it must only flag "blocking issues" if it is absolutely certain of an error, ensuring that the system does not stall due to minor ambiguities or false positives. The Critic’s output is a strict JSON object containing a binary \texttt{decision} ("approve" or "revise") and a list of \texttt{blocking\_issues} or \texttt{missing\_information}, ensuring actionable feedback is passed back to the Actor in the refinement loop.

\subsection{Iterative Refinement Protocol}
\begin{algorithm}[h]
\SetAlgoLined
\DontPrintSemicolon
\caption{Zero-shot Dual-Agent PCG Refinement}
\label{alg:dual_agent}

\KwIn{User Prompt $P_{user}$, Documentation $D$, Usage Examples $E$, Max Iterations $K$}
\KwOut{Final Parameter Trajectory $S_{final}$}

\SetKwFunction{FActor}{Actor}
\SetKwFunction{FCritic}{Critic}

$Context_{actor} \leftarrow \{P_{user}, D, E\}$\;
$S_0 \leftarrow \FActor(Context_{actor})$ \tcp*[r]{Initial generation}
$i \leftarrow 0$\;

\While{$i < K$}{
    $Feedback \leftarrow \FCritic(S_i, D, E)$\;
    
    \If{$Feedback = \emptyset$}{
        \Return{$S_i$} \tcp*[r]{Valid configuration found}
    }
    
    $Context_{actor} \leftarrow \text{UpdateContext}(S_i, Feedback)$\;
    
    $S_{i+1} \leftarrow \FActor(Context_{actor})$\;
    
    $i \leftarrow i + 1$\;
}
\Return{$S_{K}$} \tcp*[r]{Return best effort if max iterations reached}

\end{algorithm}
\noindent The core of our architecture is an internal dialogic feedback loop that allows the system to resolve ambiguities and correct hallucinations progressively. The Actor receives the Critic’s feedback combined with the original User Prompt to generate a revised trajectory ($S_{i+1}$), as shown in Algorithm~\ref{alg:dual_agent}. Through successive iterations, the trajectory becomes increasingly constrained by the functional requirements provided by the Critic, ensuring convergence toward a valid execution plan without requiring gradient updates.

\paragraph{Context Management.}
To ensure the system remains within the LLM’s effective context window across multiple iterations, we implement a \textit{state-replacement strategy}. Rather than appending the entire history of the dialogue, the system maintains a fixed-size context buffer. As shown in Figure~\ref{fig:IterativeRefinement}, the context is updated in-place: the distinct ``Initial Trajectory'' block is overwritten by the ``Revised Trajectory'' at the start of each new evaluation cycle. This preserves the most relevant state information—the current best hypothesis and the latest critique—while discarding obsolete trajectories, ensuring consistent inference latency and model performance.
\section{Experiments}
In this section, we evaluate the performance of our dual-agent Actor–Critic architecture against single-agent baselines. We focus on two key dimensions: (1) \textbf{Reliability}, assessing the system's ability to satisfy complex, multi-constraint prompt requirements; and (2) \textbf{Efficiency}, analyzing the trade-off between computational overhead and system autonomy. Our results indicate that the Actor–Critic framework not only improves task success rates by 20\% but also enhances operational efficiency, reducing the need for human follow-up prompts and decreasing overall token usage by 12.7\%.
\subsection{Experimental Setup}
\begin{figure}[htp]
    \centering
    \includegraphics[width=15cm]{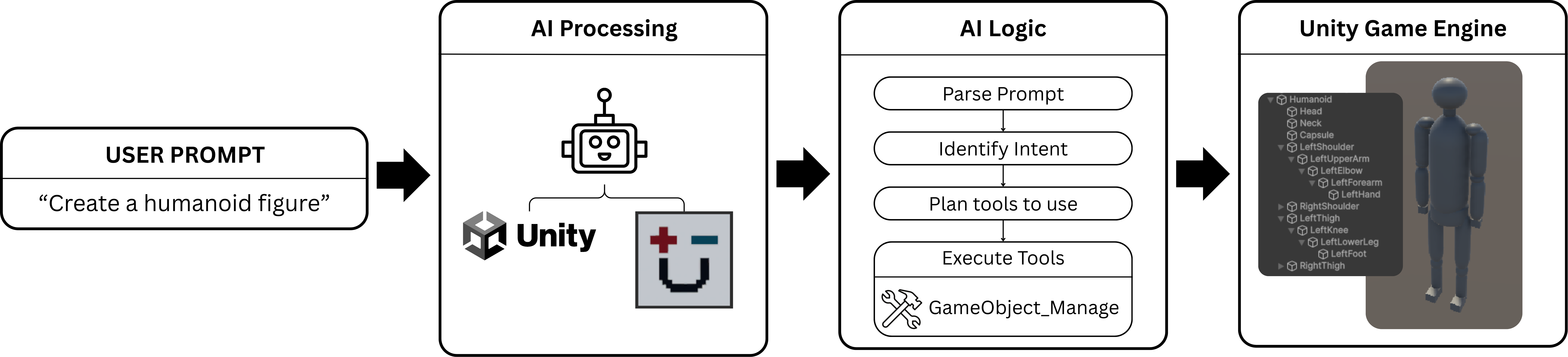}
    \caption{UGenLah system interface. The Unity-based AI assistant has access to more than 30 tools, supporting comprehensive interaction with the Unity Editor, including scene manipulation, asset management, and project configuration.}
    \label{fig:ugenlah}
\end{figure}

\paragraph{System Configuration.}
Our experiments target the \textit{TileWorldCreator} plugin within the Unity Engine. Although our architecture is tool-agnostic, we utilize this plugin to test procedural content generation (PCG) capabilities. Both the Actor and Critic agents are built upon \textit{UGenLah}, a Unity Editor AI assistant that interfaces with the editor via natural language commands (see Figure~\ref{fig:ugenlah}). We employ the \textit{Claude 4.5 Sonnet} model via API for inference. To balance creativity with instruction adherence, the Actor's temperature is set to $0.4$. The Critic's temperature is set to $0.2$ to ensure consistent and confident feedback. The maximum number of iteration cycles is set to one for all trials.

\paragraph{Experiment I: Complex Task Protocol.}
We first evaluate the reliability of our proposed Actor–Critic architecture against a single-agent baseline equipped with external resources. Both architectures are tasked with generating a complex 3D mountain map subject to four specific constraints:
\begin{itemize}
    \item \textbf{Terrain:} Formation of a single mountain peak (preventing multiple disconnected peaks).
    \item \textbf{Morphology:} The mountain must consist of exactly three height layers.
    \item \textbf{Detailing:} Grass spots must be applied specifically to the peak layer.
    \item \textbf{Scatter:} Rocks must be scattered in areas not occupied by the mountain.
\end{itemize}
All requirements are provided in the initial zero-shot prompt. No follow-up prompts are permitted, ensuring that both systems process identical information. We conduct ten independent trials to systematically evaluate success rates and error distribution.

\begin{figure}[h]
    \centering
    \includegraphics[width=15cm]{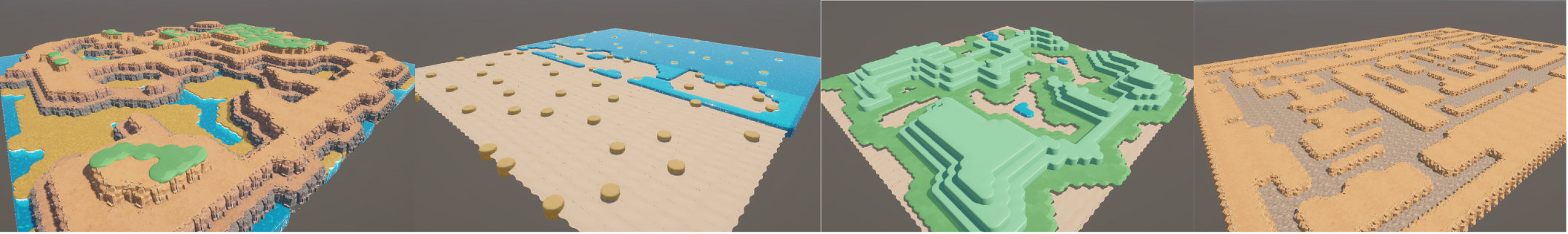}
    \caption{We evaluate the architectures on four distinct map types (two 2D and two 3D). Each map type is generated through a multi-step pipeline that applies a different combination of procedural content generation (PCG) algorithms, thereby assessing the architectures’ ability to generalize across diverse generation processes and spatial representations.}
    \label{fig:target_maps}
\end{figure}

\paragraph{Experiment II: Efficiency Protocol.}
To isolate the impact of the Critic agent, we compare our dual-agent approach against two single-agent baselines: (1) \textit{Single-Agent (w/ Docs)}, which has access to documentation and example usage, and (2) \textit{Single-Agent (No Docs)}. We evaluate these architectures across four distinct target maps (two 2D and two 3D, see Figure~\ref{fig:target_maps}), each requiring different PCG algorithms.
\\\\
Unlike Experiment I, this protocol allows for iterative refinement. If an architecture fails to produce the target map, we provide follow-up prompts describing the observed discrepancies (e.g., ``The height layers are disconnected'') without offering explicit solutions (e.g., we do not state ``reduce the y-offset''). This tests the agent's ability to self-correct based on descriptive feedback.

\paragraph{Evaluation Metrics.}
We employ distinct metrics for each experiment to assess both output quality and computational efficiency.
\begin{itemize}
    \item \textbf{Experiment I Metrics:} We report (1) \textit{Task Success Rate} and (2) \textit{Average Mistakes per Run}. A trial is classified as a \textbf{Success} if all required elements (layers, generators, modifiers) are present, even if minor parameter tuning is required. A \textbf{Failure} denotes fundamental errors that render the output unusable. We define a \textbf{Mistake} as a suboptimal parameter choice which, while not fatal to the generation process, deviates from the ideal target configuration.
    \item \textbf{Experiment II Metrics:} We track (1) \textit{Token Usage} (summed across Actor and Critic turns), (2) \textit{Follow-up Prompts Required}, and (3) \textit{Objective Completion}. An output is considered complete if the final map matches the target configuration shown in Figure~\ref{fig:target_maps} after the allowed iterations and follow up prompts.
\end{itemize}
\subsection{Results I: Reliability and Adherence}
\begin{table}[htbp]
\centering
\begin{threeparttable}
\caption{Comparison of Actor-Critic and Actor+Resources Architectures}
\label{tab:arch_comparison}
\begin{tabular}{@{}lcc@{}}
\toprule
\textbf{Metric}       & \textbf{Actor-Critic} & \textbf{Actor+Resources} \\
\midrule
Success Rate (\%)     & 80                   & 60                       \\
Average mistakes (per successful run)      & 2.25              & 2.17                  \\
\midrule
\multicolumn{3}{l}{\textbf{Common Failure Reasons\tnote{a}}} \\
\midrule
Missing Generators/Modifier\tnote{b}    & 1 (50\%)   & 2 (50\%)  \\
Missing/wrong parameter choice & 0 (0\%)   & 1 (25\%)   \\
Did not add reference layer    & 3 (100\%)   & 2 (50\%)  \\
\midrule
\multicolumn{3}{l}{\textbf{Common Mistakes (Successful Runs)}} \\
\midrule
Wrong Generator/Modifier Choice\tnote{b}    & 10 (56\%) & 4 (31\%)\\
Wrong parameter choice    & 8 (44\%) & 9 (69\%)\\
\bottomrule
\end{tabular}
\begin{tablenotes}[flushleft]
\item[a] A failed run can have more than one failure reason. Percentages represent the fraction of failed runs where each reason was observed.
\item[b] To generate maps, a Generator/Modifier must first be chosen, determining the PCG algorithm used. Parameters for the chosen Generator/Modifier can then be configured to edit the output.
\end{tablenotes}
\end{threeparttable}
\end{table}

\paragraph{Quantitative Analysis} Evaluated across ten independent trials of the 3D mountain map task, our Actor–Critic architecture achieves an 80\% task success rate, representing a 30\% relative improvement over the single-agent baseline which achieved 60\%. Although the Actor–Critic setup commits slightly more mistakes per successful run than the single-agent setup, these errors are generally non-critical and do not prevent successful map generation. By contrast, the single-agent architecture encounters critical failures more frequently, causing entire runs to fail.

\begin{figure}[h]
    \centering
    \includegraphics[width=15cm]{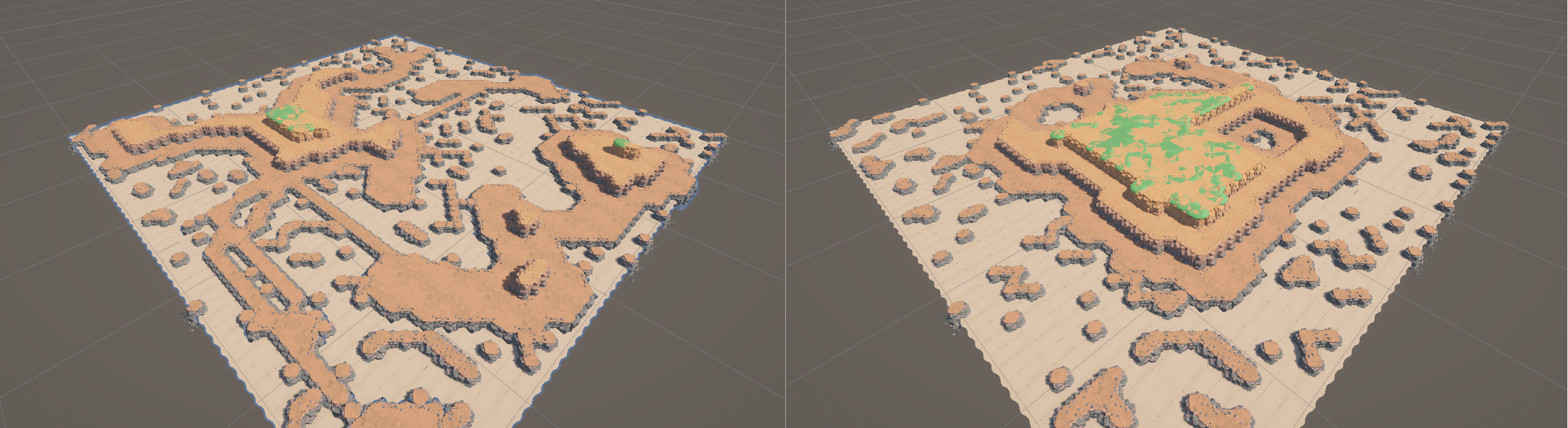}
    \caption{The actor-critic architecture showed more awareness for the requirement of creating one single mountain, configuring the parameters of Cellular Automata to create a single land mass (right) compared to connected masses that the Actor-only architecture created (left).}
    \label{fig: QualitativeComparison}
\end{figure}

\paragraph{Instruction Following} In the six successful trials, the single-agent baseline adhered to the constraint of generating a single mountain in only $50\%$ of cases. As shown in Figure~\ref{fig: QualitativeComparison}, this architecture frequently failed to parameterize the cellular automata base layer correctly, resulting in a network of disjointed land masses rather than a single mass. Conversely, the Actor–Critic architecture complied with this requirement in $88\%$ of trials. Although maps with multiple connected land masses were technically classified as successful runs, these results demonstrate that the dual-agent Actor–Critic setup is substantially more effective at interpreting user instructions and decomposing complex tasks to satisfy multiple prompt constraints.
\subsection{Results II: Impact of the Critic Agent}
\begin{table}[h]
\centering
\resizebox{\columnwidth}{!}{
    \begin{threeparttable}
    \caption{Performance comparison of our proposed architecture (Actor-Critic) against baselines.}
    \label{tab:performance_inverted}
    \begin{tabular}{@{}llccc@{}}
    \toprule
    \textbf{Map} & \textbf{Model} & \textbf{Tokens used} & \textbf{Prompts required} & \textbf{Achieves objective} \\
    \midrule
    \multirow{3}{*}{2D Beach} 
    & Actor-Critic & 16,392 & 2 & \checkmark \\
    & Actor+Resources & 18,987 & 4 & \checkmark \\
    & Actor\tnote{a} & 13,884\tnote{b} & 2\tnote{b} & \checkmark\tnote{b} \\
    \midrule
    \multirow{3}{*}{3D Mountain Island}
    & Actor-Critic & 14,583 & 4 & \checkmark \\
    & Actor+Resources & 11,873 & 5 & \checkmark \\
    & Actor & -- & -- & $\times$ \\ 
    \midrule
    \multirow{3}{*}{3D Hilly Golf Course}
    & Actor-Critic & 12,633 & 4 & \checkmark \\
    & Actor+Resources & 18,676 & 6 & \checkmark \\
    & Actor & -- & -- & $\times$ \\
    \midrule
    \multirow{3}{*}{2D Escape Maze}
    & Actor-Critic & 4,589 & 2 & \checkmark \\
    & Actor+Resources & 7,722 & 3 & \checkmark \\
    & Actor & 10,774 & 3 & \checkmark \\
    \bottomrule
    \end{tabular}
    \begin{tablenotes}[flushleft]
    \item[a] For the Actor column: “--” indicates the method was not applicable or failed to produce a result.
    \item[b] Values in this case were generated by creating C\# scripts that adjusted the TileWorldCreator output directly and not through PCG algorithms.
    \end{tablenotes}
    \end{threeparttable}
}
\end{table}
\paragraph{Operational Efficiency and Autonomy}
A comparison between the Actor–Critic architecture and the single-agent baseline (Actor+Resources) reveals that the primary advantage of the dual-agent system lies in planning quality rather than raw token economy, as shown in Table~\ref{tab:performance_inverted} While the Actor–Critic architecture achieved a modest reduction in total token usage (an average decrease of $12.7\%$ across four tasks), this metric belies a significant shift in workflow. The overhead of the Critic’s analysis was offset by a substantial reduction in the need for human intervention. The single-agent baseline required an average of $1.5$ additional follow-up prompts per task to correct ineffective strategies or guide the agent toward viable approaches. Consequently, the Actor–Critic setup proved superior not merely in efficiency, but in its ability to autonomously synthesize correct, executable plans without iterative human steering.

\paragraph{Impact of Contextual Resources}
Beyond the dual-agent dynamic, our results underscore the necessity of domain-specific context. Comparing both architectures against an off-the-shelf LLM baseline (without access to documentation or examples) highlights the significant impact of static resource injection. When supplied with documentation and usage examples, the LLM successfully grounded its generation in the specific functionalities of the \textit{TileWorldCreator} plugin and PCG principles. This domain knowledge proved sufficient for competent content generation even without complex retrieval systems (RAG). These findings suggest that for specialized software engineering tasks, the inclusion of relevant documentation is a foundational requirement that significantly enhances the model's practical utility.

\section{Generalizability and Broader Implications}

While this work utilizes Procedural Content Generation (PCG) as a testbed, the proposed dual-agent framework addresses a fundamental challenge in software engineering: the \textit{semantic gap} between high-level human intent and the rigid, high-dimensional parameter spaces of expert software.

\paragraph{Bridging the Semantic Gap in Expert Systems.}
The core innovation of our architecture is not specific to terrain generation; rather, it is a mechanism for \textit{zero-shot interface adaptation}. Traditional approaches to automating complex software rely on task-specific fine-tuning or massive datasets of user interactions. Our results demonstrate that by decoupling \textit{ideation} (Actor) from \textit{verification} (Critic), an LLM can master opaque software parameters solely through static documentation and reference examples. This suggests a scalable pathway for automating interactions with other ``expert-grade'' software—such as Computer-Aided Design (CAD) tools, scientific simulation platforms (e.g., MATLAB, Ansys), or audio synthesis engines—where parameters are often non-intuitive and the cost of trial-and-error is high.

\paragraph{Democratization of Complex Tools.}
By effectively translating natural language into executable parameter trajectories, this framework lowers the barrier to entry for professional-grade tools. Users without deep technical knowledge of specific APIs or scripting languages can leverage the full power of the underlying engine. For instance, in Digital Twin rendering or Building Information Modeling (BIM), a user could describe structural requirements in plain English, relying on the Critic to ensure the output adheres to strict engineering constraints defined in the documentation. This shifts the user's role from \textit{operator}—bogged down by syntax and UI navigation—to \textit{director}, focusing purely on high-level design goals while the agents handle the implementation details.
\section{Limitations and Future Work}

Despite the demonstrated success of the dual-agent framework in zero-shot PCG tasks, several constraints inherent to the current design warrant further investigation. While the system effectively bridges the semantic gap for single-session interactions, broader deployment would require addressing computational overhead and the lack of cross-session learning. The following subsections outline these limitations and propose avenues for future development.

\paragraph{Inference Latency and Computational Cost.}
Although the Actor–Critic architecture significantly improves generation quality, the iterative nature of the feedback loop inevitably incurs higher latency and token costs compared to single-shot baselines. The Critic's verification step doubles the inference requirement per iteration. Future work will investigate optimization strategies to mitigate this overhead, such as caching successful parameter trajectories for reuse or employing smaller, distilled models (e.g., Llama-3-8B or Phi-3) specifically fine-tuned for the Critic’s verification role, thereby reducing cost without compromising reliability.

\paragraph{Systematic Bias Analysis.}
Our experiments revealed that both architectures occasionally exhibited consistent error patterns, such as specific spatial misalignments or recurrent hallucinations of non-existent API parameters. Currently, our system corrects these biases reactively via the Critic. A key area for future work is the development of a \textit{diagnostic meta-layer} that systematically categorizes recurring failures. This would enable the transition from reactive error correction to proactive error prevention, potentially by dynamically adjusting the system prompt based on detected error trends.

\paragraph{Long-term Memory and Adaptation.}
The current system operates episodically, resetting its context buffer between tasks. Consequently, the agents cannot "learn" from previous successes or failures across different sessions. A promising direction for future research is the integration of Retrieval-Augmented Generation (RAG). By archiving successful strategies and resolved failure modes into a vector database, the system could retrieve relevant ``experiences'' for new, unseen tasks. This would allow the agents to adapt over time and improve zero-shot performance on novel maps without the need for resource-intensive parameter updates.
\section{Conclusions}
We have presented a dual-agent architecture that enables Large Language Models (LLMs) to master complex 3D procedural content generation (PCG) tasks in Unity. By decoupling generation from verification, our Actor-Critic framework achieves high-quality, zero-shot map creation, effectively bridging the semantic gap between natural language and opaque domain-specific parameters. Crucially, our results demonstrate that resource-intensive fine-tuning is not required for domain mastery; instead, providing off-the-shelf models with documentation and an iterative feedback loop is sufficient for robust tool operation. This approach significantly outperforms single-agent baselines in interpretable task decomposition and constraint satisfaction. Ultimately, this work offers a scalable and training-free paradigm for generative design, providing a flexible blueprint that can be readily transferred to other software tools and APIs.

\bibliographystyle{unsrt}
\bibliography{pcg_llm}
\end{document}